\documentclass{article}

\usepackage{iclr2027_conference,times}
\ifdefined\XeTeXversion
  \usepackage[OT1]{fontenc}
  
\fi

\usepackage{amsmath,amsfonts,bm}

\def\eqref#1{equation~\ref{#1}}

\def\1{\bm{1}}

\DeclareMathAlphabet{\mathsfit}{\encodingdefault}{\sfdefault}{m}{sl}
\SetMathAlphabet{\mathsfit}{bold}{\encodingdefault}{\sfdefault}{bx}{n}

\usepackage{graphicx}
\usepackage{booktabs}
\usepackage{placeins}
\usepackage{afterpage}
\usepackage{hyperref}
\usepackage{url}

\hypersetup{
  colorlinks=true,
  citecolor=blue,
  linkcolor=blue,
  urlcolor=blue
}

\graphicspath{{figures/}}

\title{Scaling Dexterous Manipulation via Action Representation Pretraining}

\ClearShipoutPicture
\renewcommand{\headrulewidth}{0.4pt}

\author{%
  {\bfseries Yanjinghao Xu\qquad Jiexin Zhang\qquad Huan Zhao\qquad Jianjia Mo}\\
  {\bfseries Zhenghao Yin\qquad Dingkang Liang\qquad Xiang Bai\qquad Han Ding}\\
  Huazhong University of Science and Technology\\
  Project page: \href{https://arpg-project.github.io/}{https://arpg-project.github.io/}%
}
\hypersetup{%
  pdftitle={Scaling Dexterous Manipulation via Action Representation Pretraining},
  pdfauthor={Yanjinghao Xu, Jiexin Zhang, Huan Zhao, Jianjia Mo, Zhenghao Yin, Dingkang Liang, Xiang Bai, Han Ding}%
}

\makeatletter
\def\@maketitle{\vbox{\hsize\textwidth
  {\LARGE\sc\@title\par}
  {\fontencoding{OT1}\fontfamily{ptm}\fontseries{m}\fontshape{n}\selectfont\begin{tabular}[t]{l}\rule{\z@}{24pt}\@author\end{tabular}}%
  \vskip 0.3in minus 0.1in}}
\makeatother

\begin{document}

\maketitle

\begin{abstract}
Large-scale vision, language, and video pretraining has substantially improved the perception and generalization capabilities of robotic manipulation models, but action generation for complex dexterous manipulation remains constrained by the scarcity of robot demonstrations. Limited robot demonstrations must be used to learn both action representations and task-conditioned action distributions, making it difficult to adequately learn action representations and, in turn, limiting the ability of downstream generative models to generate complex dexterous actions. To address this limitation, we propose ARPG (Action Representation Pretraining and Generation), which explicitly separates action representation learning from action generation: we first pretrain an action representation encoder on large-scale human manipulation data. We then freeze the encoder and train a compact generative backbone (130M parameters) to learn task-conditioned action distributions in the encoder's representation space. Experimental results show that ARPG achieves state-of-the-art performance on the DexJoCo benchmark and improves the average success rate on bimanual dexterous manipulation from 10.1\% for $\pi_{0.5}$ to 44.7\%. Scaling experiments further show that broader coverage of human manipulation behaviors effectively improves downstream performance. These results indicate that action representation pretraining provides an independent scaling dimension for improving complex dexterous manipulation capabilities.

\end{abstract}

\afterpage{
\begin{figure*}[t]
  \centering
  \setlength{\abovecaptionskip}{11pt}
  \begin{minipage}[c]{0.58\textwidth}
    \centering
    \IfFileExists{figures/fig1/manual/fig1_left.pdf}{%
      \includegraphics[width=\linewidth]{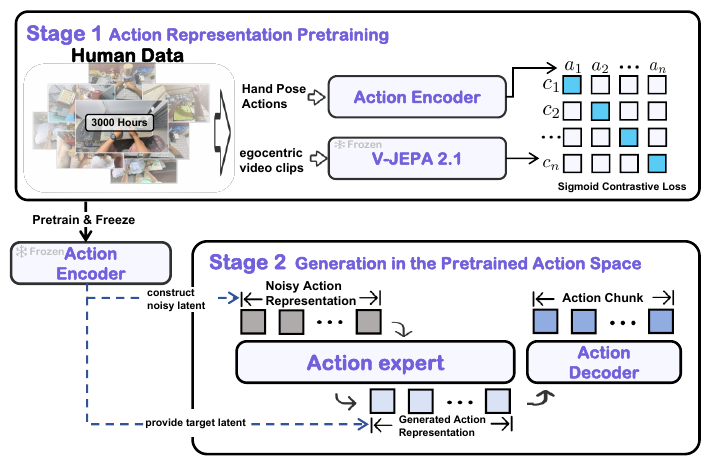}%
    }{\fbox{\parbox[c][1.65in][c]{0.94\linewidth}{\centering Manual overview pending\\\texttt{fig1\_left.pdf}}}}
  \end{minipage}\hspace{0.02\textwidth}%
  \begin{minipage}[c]{0.38\textwidth}
    \centering
    \IfFileExists{figures/fig1/fig1_right.pdf}{%
      \raisebox{-4pt}[\height][\depth]{\includegraphics[width=\linewidth]{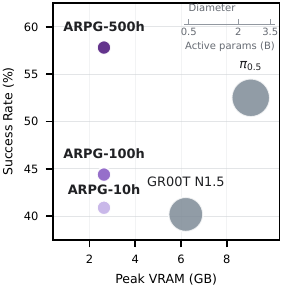}}%
    }{\fbox{\parbox[c][1.65in][c]{0.92\linewidth}{\centering Efficiency results pending\\run \texttt{make figures}}}}
  \end{minipage}
  \caption{\textbf{Overview of ARPG.} \textbf{Left:} ARPG first pretrains action representations on large-scale human manipulation data, then freezes the action encoder and trains a robot policy to generate future action-chunk representations from noisy action representations. \textbf{Right:} Comparison of average success rate and peak inference VRAM on DexJoCo; marker area indicates the number of active parameters during inference. Exact values are reported in Table~\ref{tab:dexjoco}.}
  \label{fig:overview}
  \vspace*{11pt}
\end{figure*}
}

\section{Introduction}
\label{sec:introduction}

Building general-purpose robots capable of complex dexterous manipulation, particularly coordinated and precise manipulation with dexterous hands, is an important goal in robot learning.
In recent years, large-scale vision, language, and video pretraining has endowed robot policies with strong visual representations, semantic understanding, and task generalization~\citep{bi2026motus,ye2026dreamzero,black2025pi05,zheng2026xvla}.
However, robots must ultimately translate these capabilities into concrete manipulation actions, and reliable action generation remains a significant challenge.
Mainstream policies learn action generation end to end from robot demonstrations, requiring the model to both learn representations suitable for action generation and model the action distribution under specific conditions from those demonstrations; thus, further improvements in complex action generation are constrained by both the scale of robot demonstrations and the capacity of the downstream model.
This observation raises a question: Can scalable action representation pretraining continuously improve complex dexterous manipulation without simultaneously scaling robot demonstrations and the generative backbone?

Recent research in generative modeling suggests that generative models need not learn their modeling spaces entirely from scratch: pretrained representations can either guide internal representation learning in a generative model or directly define the generative space through a pretrained encoder, thereby decoupling representation learning from subsequent distribution modeling~\citep{yu2025repa,zheng2025rae}.
However, extending this idea to robot actions is nontrivial: a representation space suitable for complex action generation must preserve sufficient action information for reliable action recovery while encoding motion and interaction semantics throughout manipulation.
Pretrained visual encoders have learned rich motion and interaction semantics from large-scale video data~\citep{assran2025vjepa2,murlabadia2026vjepa21}, but their representation spaces are not themselves designed for action representation and recovery; conversely, with action-only self-supervised learning, the action encoder must learn representations from scratch and cannot reuse the representation structure acquired through large-scale visual pretraining or exploit the scene context and interaction semantics contained in the visual modality.

Motivated by these observations, we propose ARPG (Action Representation Pretraining and Generation), which decouples action representation pretraining from downstream task-conditioned action generation and performs generation in the pretrained action representation space.
Specifically, we use a vision--action dual-tower encoder: a frozen V-JEPA~2.1~\citep{murlabadia2026vjepa21} extracts visual representations, an action Transformer encodes action trajectories, and a vision--action alignment objective based on sigmoid loss~\citep{zhai2023siglip} aligns action representations with visual representations, thereby leveraging representation priors acquired through large-scale video pretraining.
Because action representation learning is decoupled from generative learning, the action encoder need not be adapted to the action details of a specific robot embodiment through the generative objective and can instead learn reusable action representations from data; its pretraining can therefore be extended to larger-scale egocentric human manipulation data~\citep{hoque2025egodex,punamiya2026egoverse}.
After pretraining, we freeze the action tower and use it as an action representation encoder, training an action decoder on robot demonstrations to recover executable continuous actions from the representations.
On this basis, the downstream policy only needs to learn the task-conditioned action distribution in the fixed action representation space: ground-truth future action chunks~\citep{zhao2023act} are mapped by the frozen action encoder into generation targets; conditioned on the task, a DiT generative backbone~\citep{peebles2023dit} generates future action representations through Flow Matching~\citep{lipman2023flow}, which the action decoder then recovers as executable actions.

We evaluate ARPG on the DexJoCo simulation benchmark~\citep{wang2026dexjoco} and a real-robot platform.
Despite using a compact generative backbone with only approximately 130M parameters, ARPG achieves better overall performance than $\pi_{0.5}$~\citep{black2025pi05} under joint multi-task training on DexJoCo, increasing the average success rate on bimanual dexterous manipulation tasks from 10.1\% to 44.7\% while substantially reducing inference costs.
Further scaling experiments reveal that effectively scaling action representation pretraining hinges on broadening the range of task types covered by the pretraining data.
We further analyze which action representation spaces are better suited to complex action generation and verify the advantages of representations learned through vision--action alignment for downstream generation.
Thus, action representations need no longer be merely internal representations formed jointly during end-to-end policy learning; they can instead become objects of independent pretraining and scaling, providing a new scaling path for improving complex dexterous manipulation capabilities.

\section{Related Work}
\label{sec:related-work}

\textbf{Generalist Robot Policies.}
In recent years, vision--language--action (VLA) models have combined large-scale pretrained vision--language models with robot data to transfer rich visual and semantic knowledge to robot control, demonstrating strong generalization across tasks and scenes~\citep{black2025pi0,black2025pi05,zheng2026xvla}.
Recent world action models (WAMs) further leverage large-scale pretrained video models to introduce physical dynamics and interaction priors by jointly modeling future world states and actions~\citep{bi2026motus,ye2026dreamzero}.
Unlike these generalist policies, which primarily target generalization across tasks and scenes, ARPG places greater emphasis on action generation for complex dexterous manipulation and explores an alternative path to improving robotic manipulation capabilities through action representations.

\textbf{Learning from Human Manipulation Data.}
Because high-quality robot demonstrations are expensive and difficult to scale, a large body of recent work has explored using human manipulation data to scale up robot learning.
One line of work recovers hand, object, or interaction trajectories from human videos and converts them into robot-executable training data through action retargeting~\citep{yang2025egovla,zhang2026unidex,zhang2026robowheel}.
Another line of work requires no robot action annotations, instead learning latent actions or motion tokens from state changes in videos and using them as intermediate supervision connecting large-scale video pretraining and robot control~\citep{ye2025lapa,chen2025moto,luo2026jala,routray2026vipra}.
Other work directly pretrains complete models using large-scale human manipulation data, followed by adaptation with robot data~\citep{luo2026beingh0,li2025vitra,zheng2026egoscale}.
Unlike these methods, ARPG uses large-scale human manipulation data for action representation pretraining, learning reusable action representations for bimanual dexterous manipulation from rich human manipulation experience.

\textbf{Learning Action Spaces for Generation.}
Generative robot policies typically model complex multimodal action distributions directly in the raw continuous action space~\citep{chi2023diffusionpolicy}.
To reduce the difficulty of modeling high-dimensional, high-frequency action sequences, subsequent work constructs more compact action parameterizations through frequency-domain compression, vector quantization, or action autoencoders, and performs prediction and generation using discrete action tokens or learned latent action spaces~\citep{lee2024vqbet,pertsch2025fast,mete2024quest,tan2024rold}.
The above work primarily reduces the difficulty of policy modeling by compressing or reparameterizing the action space, while paying less attention to the pretrained structure of the generative space itself.
Related research in visual generation offers another perspective: classical latent-space generative methods typically learn latent spaces using reconstruction objectives~\citep{rombach2022ldm}, whereas recent methods have begun to explicitly use pretrained representations to shape the generative space, either by aligning the internal features of generative models~\citep{yu2025repa} or by directly adopting pretrained encoders to define the generative space~\citep{zheng2025rae,singh2026raev2}.
Inspired by this perspective, ARPG pretrains action representations independently and models task-conditioned action distributions for downstream tasks in a fixed pretrained representation space.

\section{Action Representation Pretraining}
\label{sec:action-representation-model}

\begin{figure*}[t]
  \centering
  \setlength{\abovecaptionskip}{11pt}
  \IfFileExists{figures/fig2/fig2_method.pdf}{%
    \includegraphics[width=0.94\textwidth]{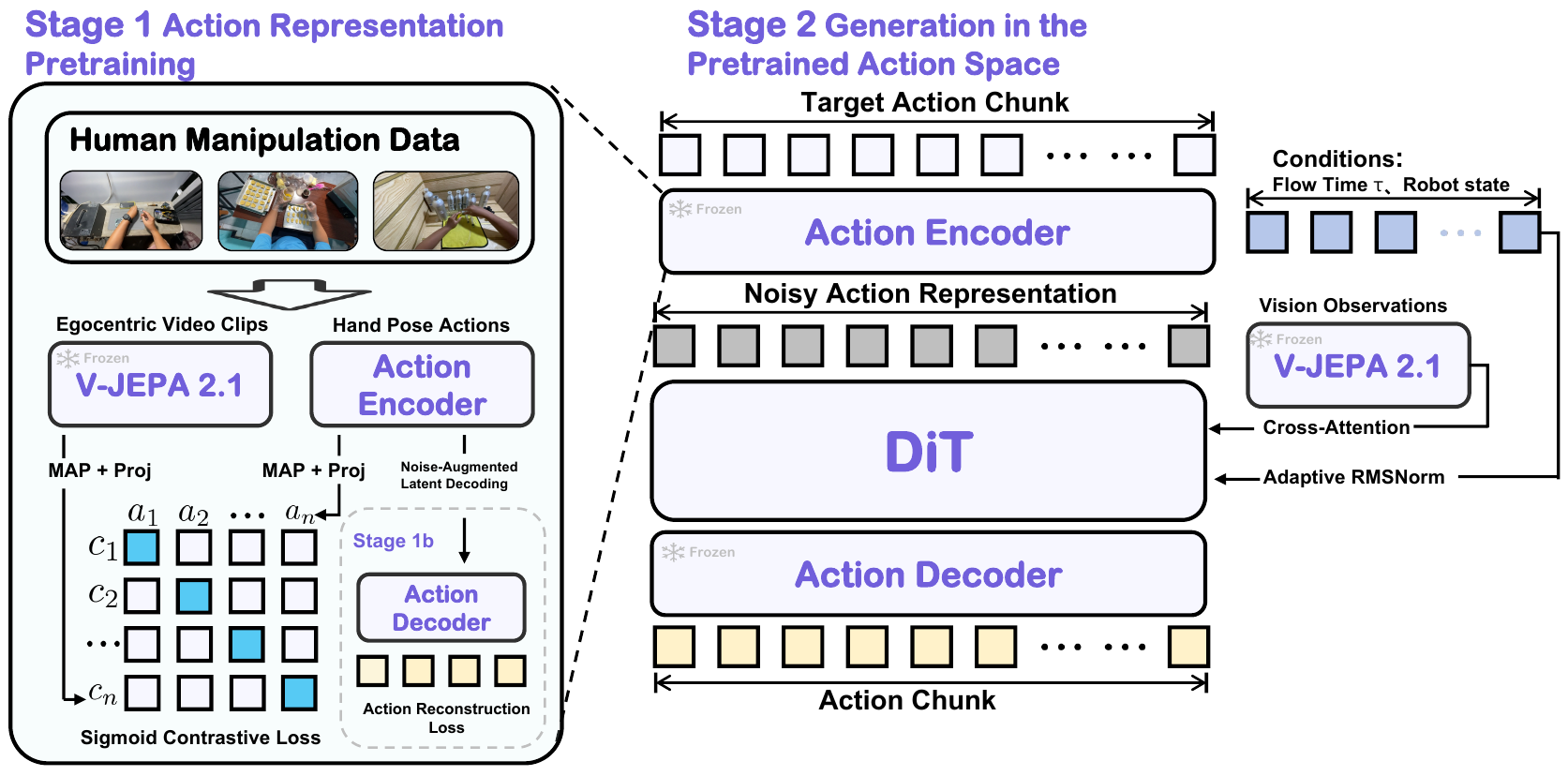}%
  }{\fbox{\parbox[c][2.0in][c]{0.96\textwidth}{\centering Manual method figure pending\\\texttt{figures/fig2/fig2\_method.pdf}}}}
  \caption{\textbf{ARPG framework.} \textbf{Stage 1a} pretrains the action encoder through vision--action alignment on human manipulation data; \textbf{Stage 1b} trains the action decoder using demonstrations from the target robot; \textbf{Stage 2} learns conditional action generation in the frozen action representation space. The action encoder is used only to construct training targets and does not participate in inference.}
  \label{fig:method}
  \vspace*{11pt}
\end{figure*}

ARPG explicitly decouples action representation learning from downstream conditional generation: it first independently learns a well-structured, decodable action representation space and then fixes this space for action generation in the subsequent stage.
The overall pipeline is shown in Figure~\ref{fig:method}.

\subsection{Action Representation Learning}
\label{sec:action-representation-encoder}

\textbf{Action representation space.}
Given a continuous action sequence $A=(a_1,\ldots,a_T)$ of length $T$, where $a_t$ denotes the action at time step $t$, we learn an action encoder $E_a$ that maps it to a temporal action representation $Z_a=E_a(A)\in\mathbb{R}^{T\times d}$, where $d$ is the representation dimension.
The representation $Z_a$ defines the action space directly modeled by the downstream generative model.
We seek an action representation space that satisfies two requirements: it should be well structured to facilitate modeling complex action distributions, and it should retain sufficient motion information to enable the recovery of executable continuous actions.

Self-supervised learning from action trajectories alone requires the representation structure to be learned from scratch, whereas pretrained video models have already acquired rich motion and interaction structure from large-scale videos; however, their visual representations cannot directly serve as a decodable action space.
Therefore, rather than directly reusing visual representations, we use them as a fixed representation reference and train an independent action encoder to align with them, thereby leveraging the structural prior provided by large-scale video pretraining while preserving action information.

\textbf{Vision--action dual-tower model.}
Given a manipulation video $V$ temporally synchronized with the action sequence $A$, the visual branch uses a frozen video encoder $E_v$ to extract spatiotemporal visual features $H_v=E_v(V)$.
We instantiate $E_v$ as V-JEPA~2.1 ViT-L/16~\citep{murlabadia2026vjepa21}; it serves only as a fixed visual reference during representation pretraining and does not participate in subsequent policy inference.
The action branch uses an $L$-layer Transformer to encode continuous action sequences; let $H_a^{(\ell)}$ denote the output of layer $\ell$, where $\ell\in\{1,\ldots,L\}$.

Inspired by the multi-layer representation aggregation design of RAEv2~\citep{singh2026raev2}, we exploit complementary information at different depths by normalizing the features from the last $K$ action layers separately and aggregating them with equal weights as $Z_a=\frac{1}{K}\sum_{\ell=L-K+1}^{L}\operatorname{Norm}(H_a^{(\ell)})$, where $K$ is the number of layers included in the aggregation and $\operatorname{Norm}(\cdot)$ denotes a shared output LayerNorm.
The resulting $Z_a$ serves both as the input to the action decoder and as the action space for the subsequent generative model.
The encoder architecture and network configuration are detailed in Appendix~\ref{app:architecture}.

\textbf{Vision--action representation alignment.}
Following the cross-modal representation alignment design of SigLIP~\citep{zhai2023siglip}, we apply it to synchronized video--action data.
For sample $i$, the visual features $H_v^i$ and action representation $Z_a^i$ each pass through Multi-head Attention Pooling (MAP) and an independent projection head, and are then normalized to obtain the global visual representation $v_i$ and action representation $u_i$.
For a mini-batch of $B$ video--action samples, $(v_i,u_i)$ from the same clip forms a positive pair, while all other cross-sample combinations form negative pairs.
We define the matching score as $s_{ij}=\tau v_i^\top u_j+b$, where $\tau$ and $b$ are learnable scale and bias parameters, respectively; let $y_{ij}=1$ when $i=j$ and $y_{ij}=-1$ otherwise.
We optimize the pairwise sigmoid alignment loss
\begin{equation}
  \mathcal{L}_{\mathrm{align}}
  =-\frac{1}{B}\sum_{i=1}^{B}\sum_{j=1}^{B}
  \log\sigma\!\left(y_{ij}s_{ij}\right),
  \label{eq:sigmoid-alignment}
\end{equation}
where $\sigma(\cdot)$ is the sigmoid function.
During training, only the action encoder $E_a$, the MAP modules on both branches, and the projection heads are updated, while the visual encoder $E_v$ remains frozen; after pretraining, only $E_a$ is retained for subsequent action modeling.
The complete training configuration is provided in Appendix~\ref{app:training}.

\subsection{Pretraining on Human Manipulation Data}
\label{sec:self-supervised-action-representation-learning}

\textbf{Cross-embodiment action representation.}
In ARPG, action representation learning is decoupled from downstream conditional generation, so the action encoder $E_a$ need not fit the control details of a specific robot embodiment through the generative objective of a robot policy and can instead focus on reusable motion and interaction structure across manipulation data.
Accordingly, action representation pretraining is no longer limited to robot demonstrations and can instead use larger-scale egocentric human manipulation data with broader behavioral coverage and greater scalability to learn the action space; demonstrations from the target robot are then used in subsequent stages for action decoding and conditional generation.

\textbf{Unified action representation.}
To allow human manipulation data from different sources and subsequent robot data to be processed by the same action encoder, we convert action trajectories into a fixed-dimensional bimanual action representation.
Specifically, the action representation at time step $t$ is $a_t=[a_t^L;a_t^R]\in\mathbb{R}^{324}$, where $a_t^L$ and $a_t^R$ denote the 162-dimensional action slots for the left and right hands, respectively.
Each action slot contains joint rotations, palm rotation, absolute wrist position, and hand keypoint positions relative to the wrist.
The redundant description of hand pose through joint rotations and keypoint positions can provide the model with implicit information about the hand's kinematic structure.
Missing or unavailable hand dimensions are handled with a validity mask.
This unified representation allows human hand trajectories and actions from different robot embodiments to be mapped to the same data interface without changing the input format of the action encoder.
The complete dimension definitions, action conversion, and adaptations for different dexterous hands are given in Appendix~\ref{app:unified-action}.

\textbf{Human-data pretraining.}
We construct temporally synchronized one-second vision--action clips from human manipulation data: the visual stream is sampled at $4\,\mathrm{Hz}$ to obtain four egocentric frames, and the action stream is sampled at $30\,\mathrm{Hz}$ to obtain an action sequence of length $T=30$.
The synchronized video $V$ and action sequence $A$ are fed into the frozen visual encoder $E_v$ and the trainable action encoder $E_a$, respectively, for pretraining with the vision--action alignment objective defined in Section~\ref{sec:action-representation-encoder}.
The pretraining data are drawn primarily from EgoDex~\citep{hoque2025egodex} and EgoVerse~\citep{punamiya2026egoverse}; data processing, the construction of different pretraining scales, and optimization configurations are detailed in Appendices~\ref{app:human-data} and~\ref{app:training}, respectively.

\subsection{Action Decoder}
\label{sec:action-decoder}

\textbf{Robot action decoding.}
After action representation pretraining, we freeze the encoder $E_a$ and train an action decoder $D_a$ using demonstrations from the target robot.
For a robot action sequence $A$, the encoder yields $Z_a=E_a(A)$, and the decoder then predicts executable actions $\hat A_{\mathrm{exec}}=D_a(Z_a)$.
The decoder is optimized with a Smooth-$L_1$ reconstruction loss using a validity mask.
The unified action format and executable action definition are given in Appendix~\ref{app:unified-action}, and the decoder architecture and training configuration are given in Appendices~\ref{app:architecture} and~\ref{app:training}.

\textbf{Noise-augmented decoding.}
Following the noise-augmented decoding design introduced in RAE~\citep{zheng2025rae}, we apply local perturbations to $Z_a$ during training and use the same executable actions as supervision to mitigate the distribution shift between generated action representations and real encoded representations, thereby improving the decoder's robustness to generation errors.
The specific perturbation scheme and training configuration are given in Appendix~\ref{app:training}.

\section{Generation in the Pretrained Action Space}
\label{sec:latent-action-generation}

After obtaining the pretrained action representation space in Section~\ref{sec:action-representation-model}, ARPG learns a conditional action distribution in this fixed space instead of directly generating raw actions. During training, both the action encoder $E_a$ and the action decoder $D_a$ remain frozen; they are used to construct generation targets and recover executable actions, respectively.

\textbf{Action Representation Targets.} At time $t$, let $A_t=(a_t,\ldots,a_{t+H-1})$ denote a future action chunk, $O_t$ the visual observation, and $q_t$ the robot state. The frozen action encoder first maps the action chunk into the pretrained action space, and normalization yields the generation target $\bar Z_t=N_z(E_a(A_t))$.
Thus, ARPG learns the conditional distribution $p_\theta(\bar Z_t\mid O_t,q_t)$ rather than directly modeling $p(A_t\mid O_t,q_t)$ in the raw control space. After denormalization, the frozen action decoder $D_a$ recovers continuous actions from the generated result.

\textbf{Conditional Representation Generation.} We use DiT~\citep{peebles2023dit} as the generative backbone and train it with Flow Matching~\citep{lipman2023flow}. The frozen V-JEPA 2.1~\citep{murlabadia2026vjepa21} extracts the visual condition $C_t^v$, and the robot state forms the state condition $C_t^s$. Given Gaussian noise $\epsilon\sim\mathcal N(0,I)$ and flow time $\tau\in[0,1]$, we construct the interpolated representation $Z_t^\tau=(1-\tau)\bar Z_t+\tau\epsilon$ and optimize
\begin{equation}
\mathcal{L}_{\mathrm{FM}}
=\mathbb{E}\!\left[
\left\|G_\theta(Z_t^\tau,\tau;C_t^v,C_t^s)
-(\epsilon-\bar Z_t)\right\|_2^2
\right].
\label{eq:flow-matching}
\end{equation}

\textbf{Training Objective.} In addition to $\mathcal L_{\mathrm{FM}}$, we introduce representation-level and action-level auxiliary supervision. Following REPA~\citep{yu2025repa}, we align the intermediate DiT feature $H_t^{(m)}$ with the target representation produced by the frozen action encoder and define $\mathcal L_{\mathrm{REPA}}=\mathbb E[\|P_\phi(H_t^{(m)})-\operatorname{sg}(\bar Z_t)\|_2^2]$, where $P_\phi$ is a projection head and $\operatorname{sg}(\cdot)$ denotes stop-gradient. In addition, $\mathcal L_{\mathrm{dec}}$ constrains the decoded action values, while $\mathcal L_{\mathrm{smooth}}$ constrains changes between adjacent actions. The final training objective is
\begin{equation}
\mathcal L
=\lambda_{\mathrm{FM}}\mathcal L_{\mathrm{FM}}
+\lambda_{\mathrm{REPA}}\mathcal L_{\mathrm{REPA}}
+\lambda_{\mathrm{dec}}\mathcal L_{\mathrm{dec}}
+\lambda_{\mathrm{smooth}}\mathcal L_{\mathrm{smooth}}.
\label{eq:total-objective}
\end{equation}

\textbf{Inference.} Starting from Gaussian noise, we generate an action representation through the conditional velocity field and obtain a future action chunk through denormalization and the frozen action decoder $D_a$. The action encoder $E_a$ is used only to construct supervision targets during training and does not participate in inference. The solver and inference settings are detailed in Appendix~\ref{app:training}.

\section{Experiments}
\label{sec:experiments}

In this section, we conduct extensive experiments focusing on the following questions: \textbf{Q1.} Does ARPG improve complex dexterous manipulation, particularly bimanual coordination, while retaining a compact and efficient downstream policy? (Table~\ref{tab:dexjoco}, Figure~\ref{fig:real_robot}) \textbf{Q2.} How should action representation pretraining be scaled to improve downstream action generation? (Figure~\ref{fig:scaling}, Table~\ref{tab:dexjoco}) \textbf{Q3.} What representation and generative modeling choices are critical for effective conditional action generation in the pretrained action space? (Table~\ref{tab:ablations})

\subsection{Experimental Setups}
\label{sec:experimental-setups}

We evaluate ARPG on a simulation benchmark and a real-world robot platform. Complete training and evaluation details are provided in Appendices~\ref{app:training}, \ref{app:simulation}, and~\ref{app:real-robot}.

\textbf{Simulation benchmark.}
We adopt DexJoCo~\citep{wang2026dexjoco} as our primary simulation benchmark. DexJoCo contains 11 task-level dexterous manipulation tasks, including six single-arm tasks and five bimanual coordination tasks. Following its original task definitions and success criteria, we report per-task success rates and compute macro-average success rates across all tasks, the single-arm tasks, and the bimanual tasks.

\textbf{Real-world robot evaluation.}
We further evaluate ARPG on three long-horizon dexterous manipulation tasks on a real-world robot platform: hammer handover, pouring coffee beans, and cabinet retrieval. Each task is divided into three ordered stages to characterize stage-wise policy performance over long-horizon execution. ARPG and $\pi_{0.5}$ are evaluated using the same robot demonstrations, initial-state distributions, and success criteria.

\textbf{Baselines.}
We compare against two representative large-scale generalist robot policies: $\pi_{0.5}$~\citep{black2025pi05} and GR00T N1.5~\citep{bjorck2025gr00tn15}. $\pi_{0.5}$ represents the current large-scale VLA paradigm and has shown strong performance in open-world generalization and long-horizon dexterous manipulation; GR00T N1.5 is a large-scale foundation model for general-purpose humanoid robots that supports bimanual and dexterous-hand manipulation and exhibits strong generalization across robot embodiments.

\subsection{Effective and Efficient Dexterous Manipulation}
\label{sec:overall-dexterous-performance}

\begin{figure*}[t]
  \centering
  \setlength{\abovecaptionskip}{7pt}
  \IfFileExists{figures/fig3/fig3_real_robot_chart.pdf}{\includegraphics[width=\textwidth]{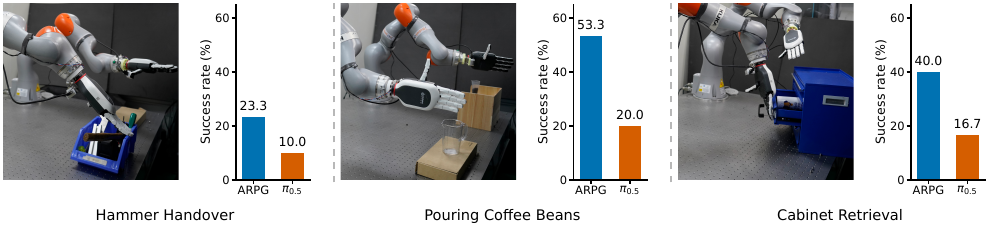}}{\fbox{\parbox[c][1.25in][c]{0.94\textwidth}{\centering Real-robot results pending}}}
  \caption{\textbf{Real-world task success rates.} We compare ARPG and $\pi_{0.5}$ on Hammer Handover, Pouring Coffee Beans, and Cabinet Retrieval. Each result is the proportion of successful rollouts over 30 trials; photographs show the task setups.}
  \label{fig:real_robot}
  \vspace*{11pt}
\end{figure*}

We first investigate \textbf{Q1}, evaluating ARPG's overall performance and deployment efficiency on complex dexterous manipulation tasks, with results shown in Table~\ref{tab:dexjoco} and Figure~\ref{fig:real_robot}.

\textbf{Dexterous manipulation performance.}
We evaluate ARPG's dexterous manipulation performance on DexJoCo, using the baseline results reported by \citet{wang2026dexjoco}. As shown in Table~\ref{tab:dexjoco}, ARPG achieves the best overall success rate under single-task training (61.3\%). Under joint multi-task training, ARPG still achieves 57.8\%, whereas $\pi_{0.5}$ drops to 35.0\%, further widening the performance gap. This advantage is particularly pronounced in bimanual dexterous manipulation: ARPG maintains a success rate of approximately 45\% under both training settings, whereas $\pi_{0.5}$ drops from 35.3\% to 10.1\%. These results indicate that the advantages of pretrained action representations are particularly pronounced in complex multi-task generation. By decoupling action representation learning from conditional generation, ARPG provides the downstream model with an action space that is easier to model, thereby alleviating the burden of learning diverse action distributions.

\begin{table*}[t]
  \centering
  \small
  \setlength{\tabcolsep}{5.0pt}
  \renewcommand{\arraystretch}{1.12}
  \caption{\textbf{Dexterous manipulation performance and deployment efficiency on the DexJoCo benchmark.} We report macro-average success rates (\%) over single-arm, bimanual, and all tasks under single-task and multi-task training, together with inference costs. Complete per-task results are in Appendix~\ref{app:simulation-evaluation}. Bold indicates the best result within each setting.}
  \label{tab:dexjoco}
  \begin{tabular*}{\textwidth}{@{\extracolsep{\fill}}lcccccc@{}}
    \toprule
    \multicolumn{1}{c}{\raisebox{-0.5\baselineskip}[0pt][0pt]{Method}} & \multicolumn{3}{c}{Inference cost} & \multicolumn{3}{c}{Success rate (\%)} \\
    \cmidrule(lr){2-4} \cmidrule(lr){5-7}
     & Params (B) $\downarrow$ & VRAM (GB) $\downarrow$ & Latency (ms) $\downarrow$ & Single arm $\uparrow$ & Bimanual $\uparrow$ & Overall $\uparrow$ \\
    \midrule
    \multicolumn{7}{@{}l}{\textit{Single-task training}} \\
    $\pi_{0.5}$ & 3.35 & 9.05 & 84.9 & 69.7 & 35.3 & 54.1 \\
    GR00T N1.5 & 2.72 & 6.21 & 93.6 & 56.2 & 21.1 & 40.2 \\
    \textbf{ARPG} & \textbf{0.46} & \textbf{2.63} & \textbf{50.8} & \textbf{75.2} & \textbf{44.5} & \textbf{61.3} \\
    \midrule
    \multicolumn{7}{@{}l}{\textit{Multi-task training}} \\
    $\pi_{0.5}$ & 3.35 & 9.05 & 84.9 & 55.8 & 10.1 & 35.0 \\
    \textbf{ARPG} & \textbf{0.46} & \textbf{2.63} & \textbf{50.8} & \textbf{68.8} & \textbf{44.7} & \textbf{57.8} \\
    \bottomrule
  \end{tabular*}
\end{table*}

\textbf{Deployment efficiency and multi-task scalability.}
At the same time, ARPG achieves this performance with substantially fewer active parameters, lower peak VRAM, and lower inference latency, demonstrating greater deployment efficiency. These results indicate that improving complex action generation need not rely solely on increasing the capacity of the generative model. This further highlights the value of independently pretraining action representations: the downstream generative model can remain compact while achieving high performance in complex action generation.

\textbf{Real-world evaluation.}
We further evaluate ARPG's real-world robot performance on three long-horizon bimanual dexterous manipulation tasks: hammer handover, pouring coffee beans, and cabinet retrieval. As shown in Figure~\ref{fig:real_robot}, under the same robot demonstrations and evaluation conditions, ARPG achieves higher full-task success rates than $\pi_{0.5}$ on all three tasks. These results show that ARPG's advantage in complex action generation observed in simulation also translates to real-world robot manipulation.

\subsection{Scaling Action Representation Pretraining for Downstream Generation}
\label{sec:scaling-pretraining}

We further investigate Q2. To this end, we scale pretraining data, action encoder size, and pretraining compute along three separate dimensions, always holding all other configurations fixed and varying only the action representation pretraining settings.

\textbf{Pretraining data scaling.}
We construct nested human manipulation datasets of 10, 100, 500, and 3,000 hours, and additionally pretrain a Robot-only encoder on 8.8 hours of robot demonstrations from DexJoCo as a control for pretraining on data from the downstream domain. As the data expand from 10 to 500 hours, the added data progressively cover more tasks and manipulation behaviors, increasing both data volume and behavioral coverage; in contrast, when the data expand from 500 to 3,000 hours, the added data are primarily concentrated in existing manipulation types, substantially increasing the number of samples for each type while providing only a very limited expansion in behavioral coverage. The detailed data composition is provided in Appendix~\ref{app:human-data}.

As shown in Figure~\ref{fig:scaling}, the downstream success rate steadily increases from 40.9\% to 57.8\% as the human pretraining data expand from 10 to 500 hours. By comparison, the encoder pretrained solely on 8.8 hours of in-domain robot data achieves a success rate of 42.7\%, which remains below that achieved with 100 hours of human pretraining data. When the human pretraining data are further increased to 3,000 hours, however, downstream performance does not improve further while behavioral coverage remains largely unchanged.

\begin{figure}[t]
  \centering
  \setlength{\abovecaptionskip}{11pt}
  \IfFileExists{figures/fig4/figure4_scaling.pdf}{\includegraphics[width=\linewidth]{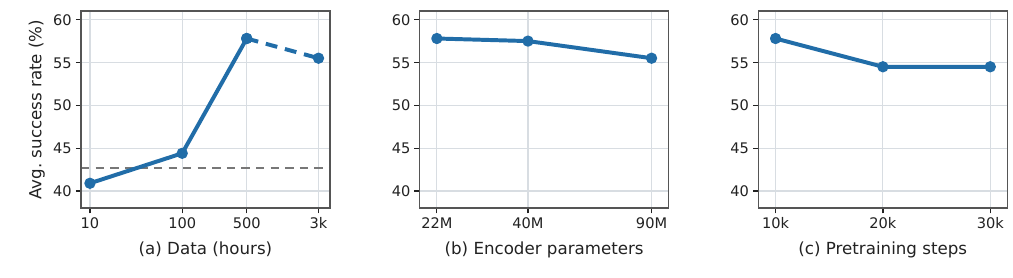}}{\fbox{\parbox[c][1.25in][c]{0.94\linewidth}{\centering Scaling results pending}}}
  \caption{\textbf{Scaling action representation pretraining.} DexJoCo average success rate as a function of (a) human pretraining data, (b) action encoder parameter count, and (c) pretraining steps. Panel (b) uses a base-2 logarithmic x-axis. In (a), the solid line connects the 10--500-hour results, the dashed blue segment connects 500 and 3,000 hours, and the dashed gray line marks the Robot-only reference.}
  \label{fig:scaling}
  \vspace*{11pt}
\end{figure}

\textbf{Behavioral coverage and diminishing returns.}
These results indicate that simply increasing the amount of pretraining data does not necessarily improve downstream performance, and that the manipulation behaviors covered by the data may also be an important factor affecting the effectiveness of action representation pretraining. We believe that this phenomenon may be attributable to the following two factors. First, our contrastive learning objective treats unpaired video--action combinations as negative samples. Consequently, when the data contain many similar manipulation behaviors, sequences with similar semantics or motion patterns may also form negative pairs, introducing potential false negatives that affect representation learning. Second, although adding samples of existing manipulation behaviors increases data volume, it may not provide new motion patterns or interaction relationships; by comparison, introducing new manipulation behaviors may expose the encoder to richer motion and interaction structures. Therefore, under the current experimental setting, expanding the behavioral coverage of pretraining data may be a scaling direction that warrants more attention than simply increasing the number of samples of existing behaviors.

\textbf{Encoder size and pretraining compute.}
Further increasing the action encoder size or pretraining compute does not consistently improve the downstream success rate. This does not imply that action representation quality has not improved; rather, it indicates that optimizing the representation pretraining objective more thoroughly does not necessarily yield action representations better suited to downstream generation. One possible explanation is that the current pretraining objective constrains sequence-level vision--action alignment after pooling, whereas the downstream DiT actually models token-level temporal action representations. Therefore, increasing model capacity or pretraining compute may further improve global representation alignment without necessarily enhancing the local temporal information on which downstream action generation relies.

\vfill

\subsection{Ablations and Analysis}
\label{sec:ablation}

Finally, we investigate \textbf{Q3}. To this end, we hold all other conditions fixed, vary only how the action representation is learned, and further ablate multilayer feature aggregation and the DiT training objective, with results shown in Table~\ref{tab:ablations}(a)--(c).

\begin{table}[t]
  \centering
  \caption{Ablations of action representation and generative design. We report the overall success rate (\%) on DexJoCo.}
  \label{tab:ablations}
  \small
  \setlength{\tabcolsep}{3.8pt}
  \renewcommand{\arraystretch}{1.08}
  \begin{tabular*}{\linewidth}{@{\extracolsep{\fill}}lr@{\hspace{1.2em}}lr@{\hspace{1.2em}}lr}
    \toprule
    \multicolumn{2}{c}{\textbf{(a) Action representation}} & \multicolumn{2}{c}{\textbf{(b) Aggregation depth}} & \multicolumn{2}{c}{\textbf{(c) Auxiliary objectives}} \\
    \cmidrule(lr){1-2} \cmidrule(lr){3-4} \cmidrule(lr){5-6}
    Representation & Overall $\uparrow$ & Last-$K$ & Overall $\uparrow$ & Objective & Overall $\uparrow$ \\
    \midrule
    Raw Action & 13.3 & 1 & 52.5 & w/o $\mathcal{L}_{\mathrm{REPA}}$ & 54.9 \\
    Action-MAE & 36.4 & 2 & \textbf{57.8} & w/o $\mathcal{L}_{\mathrm{dec}}$ & 56.0 \\
    Action-iBOT & 30.5 & 3 & 51.6 & Full & \textbf{57.8} \\
    Visual Alignment + iBOT & 45.6 &  &  &  &  \\
    Visual Alignment & \textbf{57.8} &  &  &  &  \\
    \bottomrule
  \end{tabular*}
\end{table}

\textbf{Action representation learning.}
As shown in Table~\ref{tab:ablations}(a), both Action-MAE and Action-iBOT improve over Raw Action, supporting the effectiveness of generation in a pretrained action representation space. However, the gains from learning representations solely from action trajectories remain limited. Introducing Visual Alignment further improves downstream performance and achieves the best result. We believe this improvement arises primarily because visual representations provide the action space with a structural reference shaped by large-scale pretraining: compared with action-only self-supervision, which must learn representation structure from scratch, Visual Alignment can transfer the visual model's existing representation structure and interaction semantics into the action representation while retaining the information needed for action decoding. The resulting action space is therefore easier for the downstream generative model to learn.

\textbf{Additional ablations.}
We further ablate key design choices in representation construction and downstream generative training. As shown in Table~\ref{tab:ablations}(b), when the last $K=1,2,3$ layers are used to construct the action representation, $K=2$ achieves the best result, supporting the use of multilayer deep-feature aggregation in the main model. Meanwhile, the ablations of $\mathcal{L}_{\mathrm{REPA}}$ and $\mathcal{L}_{\mathrm{dec}}$ in Table~\ref{tab:ablations}(c) show that the full DiT training objective achieves the best downstream performance, validating the effectiveness of these auxiliary constraints for generating action representations.

\FloatBarrier

\section{Limitations and Conclusion}
\label{sec:limitations-and-conclusion}

We present ARPG, which decouples action representation pretraining from downstream conditional action generation, enabling action representations to be pretrained on more readily scalable human manipulation data and providing downstream policies with a representation space better suited to generation.
Experiments show that ARPG achieves stronger dexterous manipulation performance with a compact downstream policy and demonstrates substantial advantages in multi-task bimanual manipulation.
Scaling experiments further demonstrate that broadening the coverage of manipulation behaviors in the pretraining data effectively improves downstream performance.
Overall, ARPG shows that scalable action representation pretraining offers a path to scaling complex dexterous manipulation independently of the scale of robot demonstrations and the capacity of the downstream policy.

The current work has several limitations.
First, ARPG has so far been validated only on dexterous-hand manipulation; its ability to transfer to other robot embodiments, such as grippers, remains to be studied.
Second, this work focuses primarily on complex action generation and has not systematically evaluated semantic understanding or task generalization in open environments.
Furthermore, the current pretraining data and compute scales remain limited, and scaling behavior at larger scales requires further validation.
We leave these directions for future work.

\vfill

\subsection*{AI use statement}

In this work, we used generative AI tools to assist with the preparation of the manuscript and research code. Specifically, generative AI tools were used for Chinese--English translation and language editing of technical text, assistance with manuscript wording and LaTeX editing, and software-engineering support for the research code, including code reading, refactoring, debugging, and documentation. All AI-assisted text, code, and documentation were reviewed by the authors, and all content concerning methodological descriptions, mathematical expressions, citations, and experimental results was manually verified by the authors. Generative AI tools were not used as a source of experimental data or as a substitute for simulation or real-robot experiments; all quantitative results reported in this paper were obtained from experiments executed and verified by the authors. The authors take full responsibility for the final content of this work, including any text, code, or other artifacts produced or modified with the assistance of generative AI.

\subsection*{Ethics statement}

This work uses existing large-scale egocentric human manipulation data for action representation pretraining, with the current training corpus primarily consisting of EgoDex and EgoVerse. ARPG uses manipulation videos and information related to hand motion and actions solely to learn representations for robotic manipulation; the work does not aim to perform identity recognition, biometric identification, or inference of sensitive personal attributes. For third-party human data, we follow the access, usage, and redistribution conditions specified by the respective data sources, and we do not redistribute raw human videos whose redistribution is restricted by the original licenses or access terms. Information regarding participant consent, privacy protection, and ethical approval is governed by the official documentation provided by the original dataset creators. Because egocentric videos may contain privacy-sensitive visual information, we restrict their use to manipulation representation learning within the scope permitted by the corresponding data sources.

This work also includes dexterous manipulation experiments on physical robot platforms. These experiments are conducted in controlled environments with operator supervision, emergency-stop mechanisms, robot joint and workspace constraints, motion-speed limits, and the native safety mechanisms of the robotic platforms to reduce risks associated with unintended motion, collisions, and object damage. Outputs of the learned policy are not used to bypass the existing safety layers of the robotic system. We further acknowledge that learned policies may exhibit unexpected behavior when deployed outside the robot embodiments, tasks, and environmental distributions evaluated in this work; deployment on new platforms therefore requires independent calibration, safety validation, and human supervision. The intended use of this work is research on robotic manipulation, rather than identity surveillance or applications targeting sensitive personal information.

\bibliography{references}
\bibliographystyle{iclr2027_conference}

\clearpage
\appendix
\section{Architecture Details}
\label{app:architecture}

\subsection{Action Encoder}
The action encoder $E_a$ uses a Transformer~\citep{vaswani2017attention} to model the temporal structure of continuous bimanual action sequences. For an action clip $A\in\mathbb{R}^{30\times324}$ of length $T=30$, the action at each time step is first encoded in groups according to the anatomical structure of the two hands and then mapped to a frame-level action token. After multi-layer Transformer encoding, features from multiple terminal layers are aggregated into the final action representation $Z_a\in\mathbb{R}^{30\times512}$.

\textbf{Action tokenization.}
The 324-dimensional action in each frame is divided into the wrist and five fingers for the left and right hands, yielding 12 action groups in total. Each action group is first mapped to a 32-dimensional local feature by an independent linear layer with GELU activation; the 12 group features are then concatenated and further projected into a 512-dimensional action token. Invalid action dimensions are zeroed before tokenization; when an entire time step contains no valid action, the corresponding token is masked in temporal attention.

\textbf{Transformer encoder.}
The action encoder contains eight Transformer blocks with a hidden dimension of 512, eight attention heads, an MLP ratio of 4, dropout of 0.1, and rotary positional embeddings~\citep{su2024roformer} for temporal position encoding.

\textbf{Multi-layer action representation.}
To exploit complementary action information in deep Transformer layers, we apply a shared LayerNorm to the outputs of the last $K$ layers and aggregate them uniformly along network depth:
\begin{equation}
  E_a(A)=\frac{1}{K}\sum_{\ell=L-K+1}^{L}
  \operatorname{Norm}\!\left(H_a^{(\ell)}\right).
  \label{eq:appendix-multilayer-action}
\end{equation}
The main model uses $K=2$, namely, it aggregates the outputs of the last two Transformer blocks, ultimately producing $Z_a\in\mathbb{R}^{30\times512}$.

\textbf{Sequence pooling and projection.}
During action-representation pretraining, the visual and action branches use separate Multi-head Attention Pooling (MAP) modules to aggregate token sequences into sequence-level representations. Each MAP module uses a learnable query, applies LayerNorm separately to the query and input tokens, and performs sequence aggregation through eight-head cross-attention. The 1024-dimensional pooled feature on the visual side and the 512-dimensional pooled feature on the action side are mapped into a shared 512-dimensional space by separate linear projection heads and then $\ell_2$-normalized.

\subsection{Action Encoder Training Objectives}
Given a time-synchronized first-person video clip $V_i$ and action sequence $A_i$, the frozen V-JEPA 2.1 encoder~\citep{murlabadia2026vjepa21} produces visual tokens, while the action encoder produces the corresponding action tokens. The two branches are passed through their respective MAP modules and projection heads to obtain normalized sequence-level representations $v_i$ and $a_i$. The main model in this paper uses Visual Alignment as the action-representation pretraining objective.

\textbf{Visual--action alignment.}
For a mini-batch containing $B$ synchronized video--action sequence pairs, the visual and action representations from the same clip form a positive pair, while all other cross-sequence combinations form negative pairs. The matching score between visual representation $v_i$ and action representation $a_j$ is
\begin{equation}
  s_{ij}=\tau v_i^\top a_j+b,
  \label{eq:appendix-visual-action-score}
\end{equation}
where $\tau$ and $b$ are learnable parameters. The action encoder learns an action structure consistent with the visual interaction process through a pairwise sigmoid objective~\citep{zhai2023siglip}. During training, V-JEPA 2.1 remains frozen, and only the action encoder, MAP modules, and projection heads are updated.

\textbf{Action-MAE.}
Action-MAE~\citep{he2022mae} uses the same action tokenization and Transformer encoder as the main method but does not use paired video supervision. During training, valid action groups are randomly masked, and a training-only reconstruction decoder predicts the masked raw actions. This decoder concatenates the 30 encoded latent tokens with 30 learnable query tokens carrying temporal embeddings, processes them with four pre-norm Transformer layers of hidden dimension 512 and eight attention heads, and applies LayerNorm and a linear layer to output a $30\times324$ action sequence; reconstruction loss is applied only to valid masked action dimensions. The reconstruction objective consists of a Smooth-$L_1$ reconstruction loss and a temporal velocity loss to learn local structure and temporal variation within action sequences.

\textbf{Action-iBOT.}
Action-iBOT likewise keeps the action tokenizer and Transformer encoder unchanged and uses only action-side masked self-distillation~\citep{zhou2022ibot}. During training, 50\% of the valid action content is masked, and teacher--student prediction is used to learn representations of masked action tokens. The projection head uses a 2,048-dimensional hidden layer and a 256-dimensional bottleneck and predicts 8,192 prototypes; its teacher update, temperature schedule, and other training configurations are provided in Appendix~\ref{app:training}.

\textbf{Visual Alignment.}
Visual Alignment uses only visual--action sigmoid alignment and does not introduce action-side masked self-distillation. This setting uses the motion and interaction semantic constraints provided by synchronized human-manipulation videos to structure the action-representation space.

\textbf{Visual Alignment + iBOT.}
Visual Alignment + iBOT jointly uses visual--action alignment and action-side masked self-distillation, retaining the self-supervised learning signal within action sequences while exploiting cross-modal interaction information.

\textbf{Raw Action.}
Raw Action does not use a pretrained action encoder or action decoder; instead, the generative model directly predicts continuous actions in the robot-executable action space, enabling comparison between conditional action generation in the raw action space and in the pretrained representation space.

\subsection{Action Decoder}
The action decoder $D_a$ maps the latent sequence produced by the action encoder back to the continuous actions required by the robot-control interface. Its input is $Z_a\in\mathbb{R}^{30\times512}$, and its output is a bimanual executable action sequence of the same length, $\hat A_{\mathrm{exec}}\in\mathbb{R}^{30\times198}$.

\textbf{Latent memory and query tokens.}
For the 30 latent tokens, the decoder introduces 30 learnable query tokens with temporal indices and concatenates the latent memory with the query tokens. Token-type embeddings and temporal embeddings are added separately to the two token types.

\textbf{Transformer decoder.}
The concatenated sequence is processed by a five-layer pre-norm Transformer. The decoder has a hidden dimension of 512, eight attention heads, an MLP ratio of 4, and dropout of 0.1.

\textbf{Action output.}
Only the query-token features are retained at the output and passed through LayerNorm and a linear prediction head to produce a $[30,198]$ continuous action sequence. The 198-dimensional output concatenates the 99-dimensional executable action of each hand:
\begin{equation}
  \hat A_{\mathrm{exec}}
  =[\hat A^L_{0:99};\hat A^R_{0:99}]
  \in\mathbb{R}^{30\times198}.
  \label{eq:appendix-decoder-output}
\end{equation}
The additional 126 dimensions of wrist-relative keypoint positions in the unified action representation are not output by the action decoder; their precise definition is provided in Appendix~\ref{app:unified-action}.

\subsection{Generative Backbone}
The downstream policy uses a conditional DiT~\citep{peebles2023dit} to generate future action latents in the representation space defined by the frozen action encoder. The action latent sequence serves as the main sequence of the generative model, multi-view visual tokens are injected into the generative backbone through cross-attention, and the robot proprioceptive state and flow time modulate the action features as global conditions.

\textbf{DiT backbone.}
The generative backbone contains ten Transformer blocks with a hidden dimension of 768, 12 attention heads, an MLP ratio of 4, bidirectional self-attention, temporal rotary positional embeddings, and zero dropout. The input and output projections transform between 512-dimensional action latents and 768-dimensional DiT hidden features. The main sequence contains 30 latent tokens, corresponding to a one-second, 30-step future action chunk.

\textbf{Visual conditioning.}
ARPG uses three views: a first-person camera and left and right wrist cameras. Each view provides two consecutive $384\times384$ RGB frames, which are encoded by a frozen V-JEPA 2.1 ViT-L/16. With a tubelet size of 2 and a spatial token grid of $24\times24$, each view yields 576 visual tokens of dimension 1024. Separate learnable camera embeddings are added to the three fixed views; the visual tokens are then concatenated and injected into the action latent sequence through cross-attention in every DiT block.

\textbf{Robot-state conditioning.}
The 324-dimensional robot proprioceptive state is first mapped to a 768-dimensional conditioning feature by a three-layer GELU MLP.

\textbf{Flow-time conditioning.}
The Flow Matching time $\tau$ is encoded with a sinusoidal embedding and jointly fed with the robot-state feature into a two-layer SiLU modulation network. For every DiT block, the modulation network produces the shift, scale, and residual gate required by adaptive RMSNorm, thereby injecting the current robot state and flow time into the self-attention and feed-forward sublayers.

\textbf{Intermediate representation alignment.}
REPA supervision~\citep{yu2025repa} is applied to the intermediate feature from the fourth DiT block. After RMSNorm, this feature is mapped to 512 dimensions through a linear projection head and aligned with the clean action representation produced by the frozen action encoder. Its loss weight and other auxiliary-supervision configurations are provided in Appendix~\ref{app:training}.

\FloatBarrier
\section{Unified Action Format}
\label{app:unified-action}

To unify the action forms of human-manipulation data and different robot platforms, we convert all action trajectories into a fixed-dimensional bimanual action format $A\in\mathbb{R}^{T\times324}$. Each hand corresponds to a 162-dimensional action slot consisting of local joint rotations, palm rotation, absolute wrist position, and wrist-relative keypoints. All rotations use a continuous 6D representation~\citep{zhou2019continuity}, and positions are measured in meters; hand keypoints are represented relative to the wrist so that global translation information is concentrated in the absolute wrist position. The left and right hands use the same action layout, and a validity mask handles missing or unavailable action dimensions.

\textbf{Action format.}
The 162-dimensional action for one hand consists of 15 local joint rotations, palm rotation, absolute wrist position, and 21 wrist-relative keypoints. The two 162-dimensional slots for the left and right hands are concatenated to form the complete 324-dimensional action input. The specific layout of each component is shown below, where indices use zero-based, half-open intervals.

\begin{table}[t]
  \caption{Detailed layout of the unified action format. Indices are zero-based and half-open.}
  \label{tab:d324-layout}
  \centering
  \small
  \begin{tabular}{@{}lp{0.34\textwidth}ccc@{}}
    \toprule
    Hand & Action component & Index range & Dim. & Decoder output \\
    \midrule
    Left & 15 local joint rotations (6D) & $[0,90)$ & 90 & Yes \\
    Left & Palm rotation (6D) & $[90,96)$ & 6 & Yes \\
    Left & Absolute wrist position & $[96,99)$ & 3 & Yes \\
    Left & 21 wrist-relative keypoints & $[99,162)$ & 63 & No \\
    Right & 15 local joint rotations (6D) & $[162,252)$ & 90 & Yes \\
    Right & Palm rotation (6D) & $[252,258)$ & 6 & Yes \\
    Right & Absolute wrist position & $[258,261)$ & 3 & Yes \\
    Right & 21 wrist-relative keypoints & $[261,324)$ & 63 & No \\
    \bottomrule
  \end{tabular}
\end{table}

\textbf{Executable actions.}
The complete 324-dimensional action is input to the action encoder to learn an action representation containing both kinematic state and local hand geometry. The final 63 wrist-relative keypoint dimensions of each hand serve only as auxiliary geometric information for action-representation learning and are not direct control quantities required by the robot-control interface. Therefore, the action decoder predicts only the first 99 dimensions of each hand and concatenates the left- and right-hand outputs as
\begin{equation}
  A_{\mathrm{exec}}=[A^L_{0:99};A^R_{0:99}]
  \in\mathbb{R}^{T\times198}.
  \label{eq:appendix-executable-action}
\end{equation}
Thus, the 324-dimensional unified action format is used for action encoding and representation learning, whereas the policy ultimately outputs and sends a 198-dimensional executable action to the robot controller; the additional 126 wrist-relative keypoint dimensions are not output as control commands.

\textbf{Dexterous-hand adaptation.}
To accommodate dexterous hands with different structures, we use fixed bimanual action slots and handle missing or unavailable action dimensions through a validity mask. For the four-finger Allegro hand used in DexJoCo, we use MuJoCo forward kinematics to construct the wrist-relative keypoints required by the unified action format, while dimensions corresponding to the nonexistent fifth finger are set to zero and marked invalid; these invalid dimensions are ignored when computing the relevant training losses.

\section{Human Manipulation Pretraining Data}
\label{app:human-data}

The pretraining corpus for the action encoder consists of two large-scale first-person human-manipulation datasets, EgoDex and EgoVerse, which contain time-synchronized first-person videos and 3D hand-motion trajectories. We uniformly convert the different data sources into one-second visual--action clips: the visual stream is sampled at 4 Hz to obtain four first-person images, while the action stream is sampled at 30 Hz to obtain 30 consecutive actions and converted into the 324-dimensional unified action format defined in Appendix~\ref{app:unified-action}. Human-manipulation data are used only for representation pretraining of the action encoder; the action decoder and downstream generative policy are both trained exclusively on target-robot demonstrations.

\subsection{Main Datasets}
\textbf{EgoDex.}
EgoDex~\citep{hoque2025egodex} provides first-person human-manipulation videos and 3D hand and finger motion trajectories captured with Apple Vision Pro. We use the first-person RGB videos and their corresponding hand trajectories to construct synchronized visual--action training samples, and the task labels in the dataset are not used as inputs to the action encoder.

\textbf{EgoVerse.}
EgoVerse~\citep{punamiya2026egoverse} provides geographically distributed first-person manipulation videos with hand-motion trajectories for robot learning. For clips in which only one hand is fully tracked, we retain the action trajectory of the valid hand and handle the missing second hand through the validity mask in Appendix~\ref{app:unified-action} rather than discarding the entire clip.

\subsection{Data Filtering}
The two data sources use the same action-conversion and quality-control procedure. We use a hand-tracking confidence threshold of 0.05 and require each frame to contain at least 12 valid joints. For samples whose forward-kinematics reconstruction error exceeds 0.02 m or whose inter-frame rotation change exceeds $90^\circ$, the corresponding frame is marked invalid. To reduce high-frequency jitter from hand tracking, we apply five-frame temporal smoothing within contiguous valid action intervals and avoid interpolation or smoothing across temporal intervals with missing hands. After conversion and quality control, all valid trajectories are further divided into uniform one-second training clips.

\subsection{Training-Data Scale}
We construct the 10-, 100-, 500-, and 3,000-hour pretraining sets from the same converted data pool while keeping the action format, data-preprocessing procedure, and model architecture unchanged. For the data-scaling comparison, we fix the number of pretraining optimizer steps across data scales, so the number of epochs varies with dataset size; the three epochs in Table~\ref{tab:action-representation-pretraining-hparams} apply only to the full 3,000-hour set. We order complete data shards deterministically and expand the sets cumulatively: every clip in the 10-hour set is retained in the 100-hour set, every clip in the 100-hour set is retained in the 500-hour set, and the 500-hour set is contained in the full 3,000-hour set. Each expansion adds shards without removing earlier clips; the 500-hour set approximately preserves the proportions of the two data sources in the full pool. The full set contains 11,117,819 one-second training clips. Its approximately 3,088 hours, referred to as 3,000 hours in the main paper, are the sum of clip durations rather than the duration of unique source recordings; overlapping clips, if present, are counted separately.

\begin{table}[t]
  \caption{Scale and task distribution of human-manipulation pretraining data.}
  \label{tab:human-pretraining-data}
  \centering
  \small
  \begin{tabular*}{\linewidth}{@{\extracolsep{\fill}}lrr@{}}
    \toprule
    Metric & 500 h & 3,000 h \\
    \midrule
    Summed clip duration & 500 h 20 m 16 s & 3,088 h 16 m 59 s \\
    Total samples & 1,801,216 & 11,117,819 \\
    Number of tasks & 17,382 & 23,316 \\
    \addlinespace
    Largest single-task share & 0.09\% & 9.85\% \\
    Top-10 task share & 0.77\% & 23.87\% \\
    Mean samples per task & 103.63 & 476.83 \\
    Median samples per task & 39 & 54 \\
    P90 samples per task & 213 & 453 \\
    \addlinespace
    EgoDex / EgoVerse share & 22.00\% / 78.00\% & 22.01\% / 77.99\% \\
    \bottomrule
  \end{tabular*}
\end{table}

\section{Robot Demonstrations and Downstream Data}
\label{app:robot-data}

Both the action decoder and the downstream generative policy are trained exclusively on target-robot demonstrations. For simulation experiments, we use the robot demonstrations provided by DexJoCo; for real-robot experiments, we collect data for three long-horizon manipulation tasks on a dual-arm dexterous-hand platform. Robot actions from different platforms are converted into the unified action format defined in Appendix~\ref{app:unified-action}, and the same data interface provides visual observations, robot proprioceptive states, and future action targets.

\subsection{DexJoCo Data}
DexJoCo~\citep{wang2026dexjoco} contains approximately 1.1K robot demonstrations covering 11 dexterous-manipulation tasks. We use the official \texttt{rand-obj} setting, which contains six single-arm tasks and five bimanual tasks. The simulation platform consists of Franka Panda robot arms and Wonik Allegro dexterous hands.
\begin{itemize}
  \item \textbf{Single-arm tasks:} Hammer Nail, Click Mouse, Pick Bucket, Pinch Tongs, Fold Glasses, and Water Plant.
  \item \textbf{Bimanual tasks:} Unlock iPad, Hanoi, Assembly, Microwave, and Photograph.
\end{itemize}
We follow the official DexJoCo demonstration data and environment configuration and convert the original robot trajectories into the unified training format. Each time step contains visual observations from the first-person camera and the left and right wrist cameras, the robot proprioceptive state, the action target, and the corresponding validity mask. The original TCP targets and Allegro hand-control quantities are further converted into the 324-dimensional unified action format for training the action encoder, action decoder, and downstream generative policy.

\subsection{Real-Robot Data-Collection Hardware}
The real-robot platform consists of two 7-DoF KUKA iiwa robot arms and two 12-DoF DexHand dexterous hands. Visual observations are provided by three camera views, including one Intel RealSense L515 scene camera and two Intel RealSense D405 wrist cameras mounted at the ends of the left and right robot arms. The L515 records the scene view at a resolution of $960\times540$, and the two D405 cameras record the wrist views at a resolution of $640\times480$. Robot states and multi-camera images are temporally synchronized during data recording, with a maximum permitted temporal offset of 50 ms between the robot state and visual frames.

\subsection{Real-Robot Data Content}
We collect robot demonstrations for three long-horizon bimanual dexterous-manipulation tasks: Hammer Handover, Pouring Coffee Beans, and Cabinet Retrieval. Each demonstration records synchronized three-view RGB images, dual-arm joint states, bimanual hand states, and the corresponding action targets, which are uniformly converted into the 324-dimensional action format in Appendix~\ref{app:unified-action}. The data scale for each task is as follows:

\begin{table}[t]
  \caption{Data scale for the three real-robot tasks. Durations are estimated original recording times (h:mm:ss).}
  \label{tab:real-robot-data}
  \centering
  \small
  \begin{tabular}{@{}lp{0.42\textwidth}cc@{}}
    \toprule
    Task & Task description & Demonstrations & Duration \\
    \midrule
    Hammer Handover & Pick up, hand over, and place a hammer & 40 & 3:19:31 \\
    Pouring Coffee Beans & Pick up a container and pour coffee beans & 40 & 2:15:03 \\
    Cabinet Retrieval & Open a cabinet, retrieve the target, and close it & 40 & 1:47:31 \\
    \textbf{Total} & --- & \textbf{120} & \textbf{7:22:05} \\
    \bottomrule
  \end{tabular}
\end{table}

All real-robot demonstrations are collected using hand motion-capture equipment. The action decoder and downstream policy use the same target-robot demonstrations to learn the latent-to-action mapping and the task-conditioned action-generation distribution, respectively. The durations in the table are computed from the processed training data at 30\,Hz and multiplied by three to estimate the original recording times before clipping and acceleration.

\section{Training Details}
\label{app:training}

\textbf{Action-encoder pretraining.}
The action encoder is pretrained using Visual Alignment.
The frozen V-JEPA 2.1 model provides visual representations, while only the action encoder, the visual-side and action-side MAP modules, and the projection heads are optimized during training.
For a mini-batch containing $B$ synchronized video--action clips, the visual--action pair from the same clip is treated as a positive, whereas all cross-clip combinations are treated as negatives and optimized with a pairwise sigmoid objective.
The main training hyperparameters are summarized in Table~\ref{tab:action-representation-pretraining-hparams}.

\textbf{Downstream generative-policy training.}
Both the action encoder and action decoder remain frozen throughout downstream policy training.
The Conditional DiT uses the action representations produced by the frozen action encoder as the generation targets for Flow Matching and learns the conditional distribution of future action latents given visual observations and robot proprioceptive states.
In addition to the main Flow Matching objective, we employ REPA representation alignment, decoded-action supervision, and trajectory-smoothness supervision.
The main training and inference configurations are summarized in Table~\ref{tab:downstream-policy-hparams}.

\begin{table*}[t]
  \centering
  \begin{minipage}[t]{0.485\textwidth}
    \centering
    \refstepcounter{table}\label{tab:action-representation-pretraining-hparams}
    \parbox{\linewidth}{\centering\small\textbf{Table~\thetable: Hyperparameters for action representation pretraining.}}
    \vspace{4pt}
    \scriptsize
    \renewcommand{\arraystretch}{1.05}
    \begin{tabular}{@{}ll@{}}
      \toprule
      Pretraining objective & Visual Alignment \\
      Optimizer & AdamW \\
      Batch size & 1,536 \\
      Learning rate & $2\times10^{-4}$ \\
      Minimum learning rate & $2\times10^{-6}$ \\
      Weight decay & 0.05 \\
      Warmup steps & 1,000 \\
      LR schedule & Cosine \\
      Training steps & 21,714 \\
      Training epochs & 3 \\
      Precision & bfloat16 \\
      Gradient clipping & 1.0 \\
      MAP attention heads & 8 \\
      Projection dimension & 512 \\
      Initial logit scale & 10 \\
      Initial bias & $-10$ \\
      \bottomrule
    \end{tabular}
  \end{minipage}
  \hfill
  \begin{minipage}[t]{0.485\textwidth}
    \centering
    \refstepcounter{table}\label{tab:downstream-policy-hparams}
    \parbox{\linewidth}{\centering\small\textbf{Table~\thetable: Hyperparameters for downstream policy training.}}
    \vspace{4pt}
    \scriptsize
    \renewcommand{\arraystretch}{1.05}
    \begin{tabular}{@{}ll@{}}
      \toprule
      Optimizer & AdamW \\
      Batch size & 128 \\
      Learning rate & $1\times10^{-4}$ \\
      Minimum learning rate & $1\times10^{-6}$ \\
      Weight decay & 0.05 \\
      Warmup steps & 500 \\
      LR schedule & Cosine \\
      Training steps & 120,000 \\
      Precision & bfloat16 \\
      Gradient clipping & 1.0 \\
      EMA decay & 0.9999 \\
      Flow-time sampling & $0.999\,\mathrm{Beta}(1.5,1)+0.001$ \\
      $\lambda_{\mathrm{FM}}$ & 1.0 \\
      $\lambda_{\mathrm{REPA}}$ & 0.5 \\
      $\lambda_{\mathrm{dec}}$ & 1.0 \\
      $\lambda_{\mathrm{smooth}}$ & 0.01 \\
      ODE solver & Euler \\
      Inference steps & 10 \\
      \bottomrule
    \end{tabular}
  \end{minipage}
\end{table*}

\textbf{Action-decoder training.}
After action-encoder pretraining, the action encoder remains frozen, and the action decoder is trained using only demonstrations from the target robot.
Given an action sequence $A$, the frozen encoder first produces an action latent $Z_a=E_a(A)$, after which the decoder predicts the 198-dimensional executable action defined in Appendix~\ref{app:unified-action}.
The decoder uses a Smooth-$L_1$ reconstruction loss with a validity mask, and invalid action dimensions do not contribute to the loss.

To improve the decoder's robustness to deviations in generated latents, we employ noise-augmented decoding in the standardized latent space.
For each sample, we first draw
\begin{equation}
  \sigma\sim\mathcal{U}(0,\delta),
  \qquad
  \epsilon\sim\mathcal{N}(0,\sigma^2I),
\end{equation}
and feed the perturbed latent
\begin{equation}
  \widetilde{Z}_a=Z_a+\epsilon
\end{equation}
to the decoder.
Twenty-five percent of training samples retain clean latents, and the maximum noise scale $\delta$ increases linearly from 0 to 0.05 over the first 5,000 training steps.
The decoder is trained with AdamW for 100,000 steps using a batch size of 512, a learning rate of $2\times10^{-4}$, and a weight decay of 0.05; the first 1,000 steps use linear warmup, followed by a cosine schedule that decays to $2\times10^{-6}$.
Training uses bfloat16 mixed precision and clips the gradient norm to 1.0.

\textbf{Auxiliary training objectives.}
The complete downstream-policy training objective is
\begin{equation}
  \mathcal{L}
  =
  \lambda_{\mathrm{FM}}\mathcal{L}_{\mathrm{FM}}
  + \lambda_{\mathrm{REPA}}\mathcal{L}_{\mathrm{REPA}}
  + \lambda_{\mathrm{dec}}\mathcal{L}_{\mathrm{dec}}
  + \lambda_{\mathrm{smooth}}\mathcal{L}_{\mathrm{smooth}}.
\end{equation}
Here, $\mathcal{L}_{\mathrm{REPA}}$ aligns intermediate DiT features with the clean action representations produced by the frozen action encoder; $\mathcal{L}_{\mathrm{dec}}$ directly supervises the executable actions decoded from generated latents; and $\mathcal{L}_{\mathrm{smooth}}$ constrains action changes between adjacent time steps.
All action-space losses use the corresponding validity masks; for the smoothness loss, a pair is excluded whenever an action dimension is invalid at either of the two adjacent time steps.

\FloatBarrier

\section{Simulation Experiment Details}
\label{app:simulation}

\textbf{Training setup.}
We conduct both single-task training and multi-task joint training under the \texttt{rand-obj} setting of DexJoCo.
In the single-task setting, we train a separate policy on the corresponding robot demonstrations for each task for 60,000 steps; in the multi-task setting, we jointly use data from all 11 tasks and train for 120,000 steps.
The two settings use the same model architecture and optimization configuration; the detailed training hyperparameters are provided in Appendix~\ref{app:training}.

No additional validation set is constructed during multi-task training, and the checkpoint at 120,000 steps is used for final evaluation.
Multi-task training uses a task-aware shard-local batch sampler that draws samples without replacement within each shard group and favors coverage of distinct tasks in each mini-batch; it does not enforce equal global sampling probabilities across tasks.

\textbf{Baseline implementations.}
We compare ARPG with $\pi_{0.5}$ and GR00T N1.5 as reported by DexJoCo.
To avoid additional implementation differences introduced by retraining large generalist policies, all baseline results use the values officially reported by DexJoCo under the same \texttt{rand-obj} setting~\citep{wang2026dexjoco}.
\begin{itemize}
  \item \textbf{$\pi_{0.5}$.} Both the single-task and multi-task results are taken from the official DexJoCo report.
  \item \textbf{GR00T N1.5.} The single-task results are taken from the official DexJoCo report.
  \item \textbf{ARPG.} All ARPG models are retrained using our action-format conversion and training pipeline and are evaluated separately in the single-task and multi-task settings.
\end{itemize}

\textbf{Evaluation protocol.}
All methods follow the official DexJoCo environment settings and task-success criteria.
For each task, we evaluate with random seeds 0, 1, and 2, running 50 episodes per seed for a total of 150 rollouts.
Per-task results report the mean and standard deviation of the success rates across the three random seeds.

For the single-arm tasks, bimanual tasks, and all tasks, we compute unweighted macro-averages as
\begin{equation}
  S_{\mathrm{single}} = \frac{1}{6}\sum_{i=1}^{6}S_i,\qquad
  S_{\mathrm{bimanual}} = \frac{1}{5}\sum_{i=1}^{5}S_i,\qquad
  S_{\mathrm{all}} = \frac{1}{11}\sum_{i=1}^{11}S_i.
\end{equation}
Both the policy and control frequencies are 30\,Hz.
Each policy query generates a 30-step action chunk, and the inference scheduler uses a nominal eight-step execution window for rolling replanning.
The evaluation configuration uses strict RTC (PI-VJP) execution with inference-time guidance from the previous action chunk; the prefix-attention schedule is exponential, the maximum guidance weight is 5.0, and the inference-latency safety margin is 40\,ms.

\section{Real-World Experiment Details}
\label{app:real-robot}

\textbf{Task setup.}
We evaluate the policy on three real-world robot tasks that require multiple manipulation stages to be executed sequentially: Hammer Handover, Pouring Coffee Beans, and Cabinet Retrieval.
Each task is divided into three ordered stages to evaluate both local manipulation ability and long-horizon task completion.

\begin{table*}[t]
  \centering
  \small
  \setlength{\tabcolsep}{6pt}
  \caption{Stage definitions for the real-world evaluation tasks.}
  \label{tab:real-world-task-stages}
  \begin{tabular}{@{}p{0.18\textwidth}p{0.20\textwidth}p{0.23\textwidth}p{0.25\textwidth}@{}}
    \toprule
    Task & Stage 1 & Stage 2 & Stage 3 \\
    \midrule
    Hammer Handover & Pick up the hammer & Complete the bimanual handover & Place the hammer at the target \\
    Pouring Coffee Beans & Pick up the bean container & Successfully pour out the coffee beans & Completely pour out the coffee beans \\
    Cabinet Retrieval & Open the cabinet door & Retrieve the target object & Close the cabinet door \\
    \bottomrule
  \end{tabular}
  \vspace{2pt}
  \parbox{0.96\textwidth}{\footnotesize \textit{Note:} The three stages of each task must be completed in order.}
\end{table*}

\begin{figure*}[t]
  \centering
  \includegraphics[width=\textwidth]{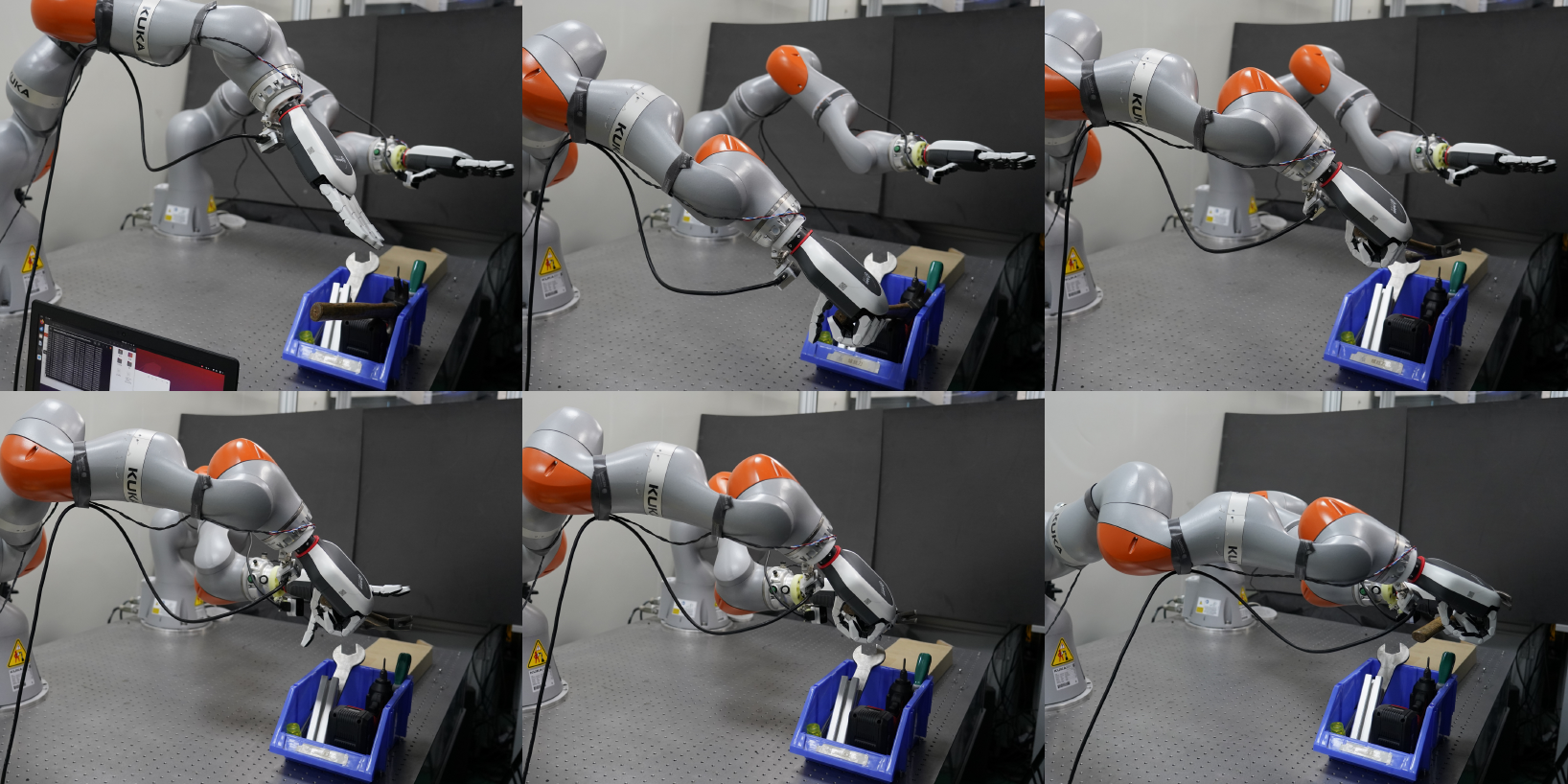}
  \caption{\textbf{Hammer Handover: Detailed Procedure.} The right hand picks up the hammer, hands it to the left hand, and the left hand puts it down.}
  \label{fig:real-hammer-process}
\end{figure*}

\begin{figure*}[t]
  \centering
  \includegraphics[width=\textwidth]{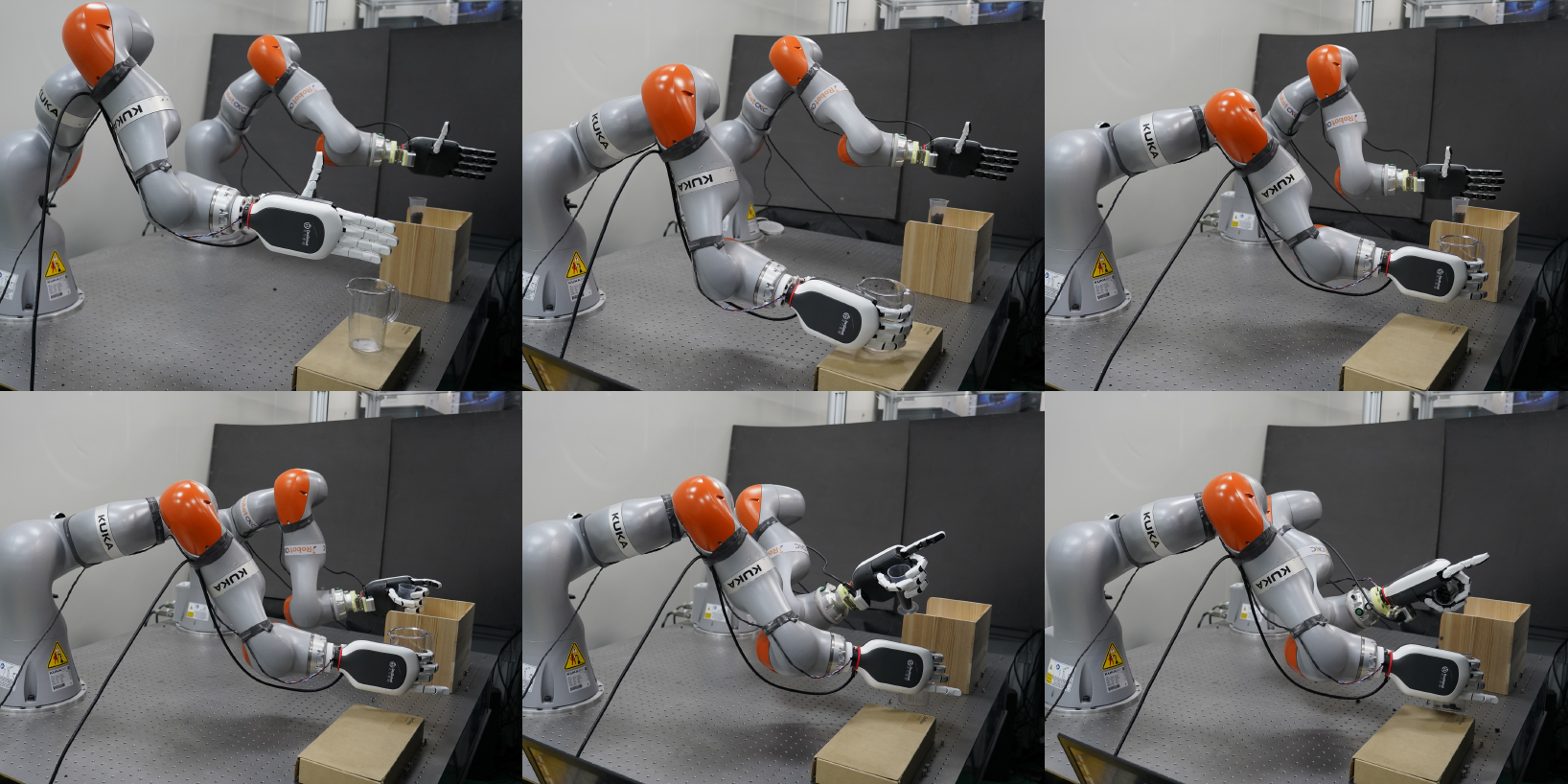}
  \caption{\textbf{Pouring Coffee Beans: Detailed Procedure.} Both hands pick up measuring cups, and the coffee beans are poured from the left-hand cup into the right-hand cup.}
  \label{fig:real-coffee-process}
\end{figure*}

\begin{figure*}[t]
  \centering
  \includegraphics[width=\textwidth]{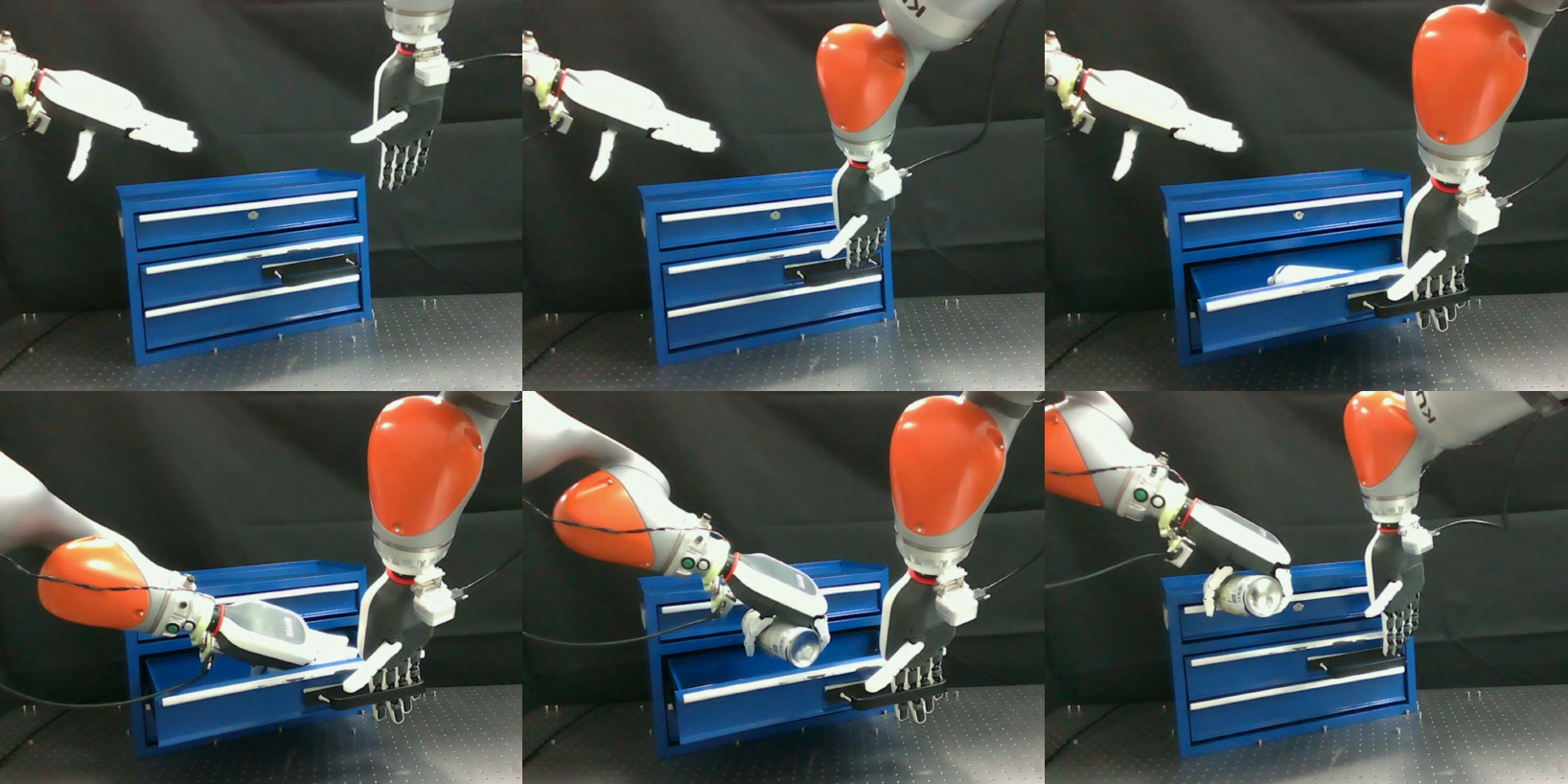}
  \caption{\textbf{Cabinet Retrieval: Detailed Procedure.} The right hand opens the cabinet, the left hand retrieves the object, and the right hand closes the cabinet.}
  \label{fig:real-cabinet-process}
\end{figure*}

The execution sequences of the three real-world tasks are shown in Figures~\ref{fig:real-hammer-process}, \ref{fig:real-coffee-process}, and~\ref{fig:real-cabinet-process}, respectively.
The six frames in each figure are ordered from left to right and then from top to bottom.

For each rollout, completion of a later stage is counted only if the preceding stage has been completed successfully; final task success requires all three stages to be completed in order.
\textbf{Evaluation protocol.}
ARPG and $\pi_{0.5}$ are trained and evaluated using the same robot demonstrations, initial-state distribution, and task-success criteria.

The success rate of each stage is computed as the proportion of rollouts that satisfy the success criterion for that stage, while the final success rate is the proportion of rollouts that complete all three stages.
For method $m$ and stage $k$, the stage success rate is defined as
\begin{equation}
  S_{m,k} = \frac{N^{\mathrm{success}}_{m,k}}{N^{\mathrm{rollout}}_m}.
\end{equation}

\textbf{Baseline setup.}
The real-world experiments primarily compare ARPG with $\pi_{0.5}$.
To ensure a consistent comparison, $\pi_{0.5}$ and ARPG use exactly the same robot demonstrations and are evaluated with the same three-view visual inputs, initial-state distribution, and success criteria.
For $\pi_{0.5}$, we initialize from the $\pi_{0.5}$ base weights and use mixed LoRA fine-tuning; the training configuration specifies 25,000 steps, a batch size of 32, a 1,000-step warmup, and checkpoint saving every 1,000 steps. The inputs comprise three-view images, task-language instructions, and a normalized 38-dimensional robot state; \texttt{discrete\_state\_input=True} discretizes the state and includes it in the language prompt, with \texttt{max\_token\_len=250}. The three task instructions describe picking up the hammer with the right hand, handing it to the left hand, and putting it down; picking up the measuring cups with both hands and pouring coffee beans from the left cup into the right cup; and opening the cabinet with the right hand, retrieving the object with the left hand, and closing the cabinet with the right hand.

\FloatBarrier

\section{Evaluation Details on Robotics Simulation}
\label{app:simulation-evaluation}

We report complete task-level success rates for all 11 DexJoCo tasks, comprising six single-arm tasks and five bimanual-coordination tasks.
All results follow the evaluation protocol described in Appendix~\ref{app:simulation}.
For experiments with multiple random seeds, we report the mean success rate and standard deviation across three seeds; we additionally report the macro-average success rates over the single-arm tasks, bimanual tasks, and all tasks.

\begin{table*}[t]
  \centering
  \scriptsize
  \setlength{\tabcolsep}{4.0pt}
  \renewcommand{\arraystretch}{1.05}
  \caption{Full task-level results on the DexJoCo benchmark.}
  \label{tab:dexjoco-full}
  \begin{tabular*}{0.98\textwidth}{@{\extracolsep{\fill}}lccccc@{}}
    \toprule
    Task & \multicolumn{3}{c}{Single-task training} & \multicolumn{2}{c}{Multi-task training} \\
    \cmidrule(lr){2-4} \cmidrule(lr){5-6}
     & $\pi_{0.5}$ & GR00T N1.5 & ARPG & $\pi_{0.5}$ & ARPG \\
    \midrule
    \multicolumn{6}{l}{\textit{Single-arm}} \\
    Hammer Nail & 84.7 $\pm$ 5.0 & 67.3 $\pm$ 4.2 & 72.0 $\pm$ 9.2 & 74.7 $\pm$ 2.3 & 52.7 $\pm$ 5.0 \\
    Click Mouse & 64.7 $\pm$ 8.1 & 85.3 $\pm$ 3.1 & 73.3 $\pm$ 8.1 & 48.0 $\pm$ 6.9 & 96.0 $\pm$ 3.3 \\
    Pick Bucket & 84.0 $\pm$ 7.2 & 72.0 $\pm$ 6.0 & 80.0 $\pm$ 2.0 & 60.7 $\pm$ 2.3 & 78.0 $\pm$ 1.6 \\
    Pinch Tongs & 24.0 $\pm$ 6.9 & 12.7 $\pm$ 2.3 & 72.7 $\pm$ 13.0 & 44.0 $\pm$ 15.9 & 50.0 $\pm$ 1.6 \\
    Fold Glasses & 72.0 $\pm$ 3.5 & 27.3 $\pm$ 2.3 & 58.0 $\pm$ 2.0 & 28.0 $\pm$ 4.0 & 41.3 $\pm$ 4.1 \\
    Water Plant & 88.7 $\pm$ 3.1 & 72.7 $\pm$ 1.2 & 95.3 $\pm$ 2.3 & 79.3 $\pm$ 9.5 & 94.7 $\pm$ 3.4 \\
    \addlinespace[2pt]
    \multicolumn{6}{l}{\textit{Bimanual}} \\
    Unlock iPad & 36.0 $\pm$ 14.4 & 12.7 $\pm$ 11.0 & 30.0 $\pm$ 3.5 & 0.0 $\pm$ 0.0 & 25.3 $\pm$ 6.6 \\
    Hanoi & 23.3 $\pm$ 7.0 & 0.7 $\pm$ 1.2 & 47.3 $\pm$ 11.5 & 4.7 $\pm$ 3.1 & 30.7 $\pm$ 3.4 \\
    Assembly & 2.7 $\pm$ 2.3 & 0.7 $\pm$ 1.2 & 8.7 $\pm$ 2.3 & 4.0 $\pm$ 5.3 & 7.3 $\pm$ 0.9 \\
    Microwave & 60.0 $\pm$ 5.3 & 50.7 $\pm$ 4.6 & 70.7 $\pm$ 6.1 & 34.7 $\pm$ 13.6 & 84.7 $\pm$ 5.3 \\
    Photograph & 54.7 $\pm$ 8.1 & 40.7 $\pm$ 7.0 & 66.0 $\pm$ 7.2 & 7.3 $\pm$ 5.0 & 75.3 $\pm$ 6.2 \\
    \midrule
    \textbf{Single-arm Avg.} & \textbf{69.7} & \textbf{56.2} & \textbf{75.2 $\pm$ 2.5} & \textbf{55.8} & \textbf{68.8} \\
    \textbf{Bimanual Avg.} & \textbf{35.3} & \textbf{21.1} & \textbf{44.5 $\pm$ 4.1} & \textbf{10.1} & \textbf{44.7} \\
    \textbf{All Avg.} & \textbf{54.1 $\pm$ 1.8} & \textbf{40.2 $\pm$ 0.3} & \textbf{61.3 $\pm$ 2.8} & \textbf{35.0 $\pm$ 2.8} & \textbf{57.8 $\pm$ 0.4} \\
    \bottomrule
  \end{tabular*}
  \vspace{2pt}
  \parbox{0.98\textwidth}{\tiny \textit{Note:} Values are success rates (\%); $\pm$ denotes the standard deviation across three seeds when available.}
\end{table*}

\end{document}